\documentclass{article}
\usepackage{fancyhdr}

\usepackage{ijcai24}

\usepackage{times}
\usepackage{soul}
\usepackage{url}
\usepackage[hidelinks]{hyperref}
\usepackage[utf8]{inputenc}
\usepackage[small]{caption}
\usepackage{graphicx}
\usepackage{amsmath}
\usepackage{amsthm}
\usepackage{booktabs}
\usepackage{algorithm}
\usepackage{algorithmic}
\usepackage[switch]{lineno}
\usepackage{amssymb}
\usepackage{amsthm}
\usepackage{bm}
\usepackage{color}
\newtheorem{definition}{Definition}
\usepackage{graphicx}

\title{FTS: A Framework to Find a Faithful TimeSieve}

\author{
Songning Lai\thanks{The first two authors contributed equally to this work.}$^{,1,3,7}$
\and
Ninghui Feng$^{*,1}$\and
Haochen Sui$^{4}$\and
Ze Ma$^{5}$\and
Hao Wang$^{6}$\and
Zichen Song$^{8}$\and \\
Hang Zhao$^{1}$\and
Yutao Yue\thanks{Correspondence to Yutao Yue \{yutaoyue@hkust-gz.edu.cn\}}$^{,1,2,3}$\\
\affiliations
$^1$The Hong Kong University of Science and Technology (Guangzhou) \\
$^2$Institute of Deep Perception Technology, JITRI \\
$^3$Deep Interdisciplinary Intelligence Lab ($DI^2 Lab$)\\
$^4$University of Michigan - Ann Arbor\\
$^5$Columbia University\\
$^6$Carnegie Mellon University\\
$^7$Shandong University\\
$^8$Lanzhou University
\emails
songninglai@hkust-gz.edu.cn,
2020303020215@neepu.edu.cn,
hcsui@umich.edu,
zm2385@columbia.edu,
haow2@alumni.cmu.edu,
songzch21@lzu.edu.cn,
hangzhao@hkust-gz.edu.cn,
yutaoyue@hkust-gz.edu.cn
}

\begin{document}

\maketitle

\begin{abstract}
The field of time series forecasting has garnered significant attention in recent years, prompting the development of advanced models like TimeSieve, which demonstrates impressive performance.
However, an analysis reveals certain unfaithfulness issues, including high sensitivity to random seeds and minute input noise perturbations. Recognizing these challenges, we embark on a quest to define the concept of \textbf{\underline{F}aithful \underline{T}ime\underline{S}ieve \underline{(FTS)}}, a model that consistently delivers reliable and robust predictions.
To address these issues, we propose a novel framework aimed at identifying and rectifying unfaithfulness in TimeSieve. Our framework is designed to enhance the model's stability and resilience, ensuring that its outputs are less susceptible to the aforementioned factors. Experimentation validates the effectiveness of our proposed framework, demonstrating improved faithfulness in the model's behavior.
Looking forward, we plan to expand our experimental scope to further validate and optimize our algorithm, ensuring comprehensive faithfulness across a wide range of scenarios. Ultimately, we aspire to make this framework can be applied to enhance the faithfulness of not just TimeSieve but also other state-of-the-art temporal methods, thereby contributing to the reliability and robustness of temporal modeling as a whole.
\end{abstract}

\section{Introduction}
\begin{figure}[tbp]
\vspace{-10pt}
  \centering
  \includegraphics[width=\linewidth]{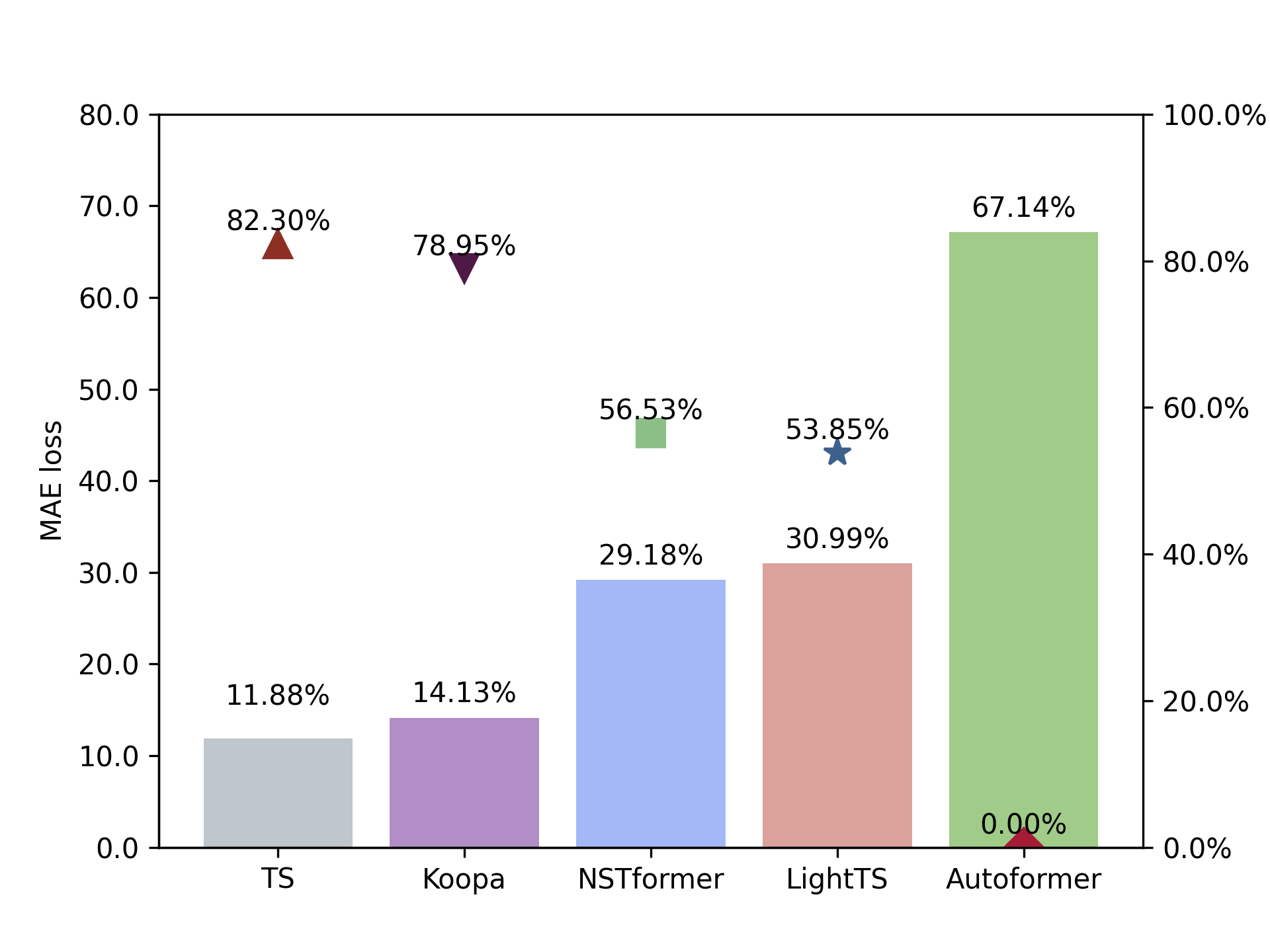}
    \vspace{-10pt}
  \caption{Visualization of the performance of the TimeSieve model versus other state-of-the-art models on the Exchange dataset. The left ordinate represents the MAE loss of the model, and the right ordinate represents the percentage of improved performance of the model over Autoformer.}
   \vspace{-10pt}
  \label{fig:sota_time}
\end{figure}

Time series forecasting is a classical learning problem that consists of analyzing time series to predict future trends based on historical information \cite{8053243,greff2016lstm}. 
In the thriving field of time series forecasting, a plethora of advanced model \cite{nie2023time,liu2023nonstationary,wu2023timesnet,yi2023frequencydomain} have emerged, pushing the boundaries of predictive capabilities.
Among these innovative approaches, we have developed TimeSieve \cite{TimeSieve_repo}, a state-of-the-art model that combines wavelet transform \cite{zhang2019wavelet,pathak2009wavelet,bentley1994wavelet} and information bottleneck theory \cite{shwartz2017opening} and embodies recent advances in the field, providing superior performance and novel insights. Figure \ref{fig:sota_time} shows the experimental results show that its performance is the best among all advanced models \cite{liu2024koopa,campos2023lightts,liu2022non,wu2021autoformer}.

Although TimeSieve exhibits powerful performance, it also demonstrates instability when facing perturbations or noise. Specifically, during our experimental process, we observed significant variability influenced by random seed perturbations, with some differences reaching up to 50\%. To show this issue, we conducted training using five random seeds, designating one of the trained models as the baseline. The test results of the remaining four models were compared with the baseline model's test results to calculate the percentage of variation. The visualization of these results is presented in Figure \ref{fig:rs}. In addition to this, we only added a small perturbation to the test set of inputs ($x(t)'=x(t) + \mathcal{N}(0,\sigma)$ with a perturbation of a certain radius $\sigma=0.1$), and the experimental results showed that the performance changed or decreased by 28.35\% (see in Table \ref{tab:rob}). These issues can seriously weaken the faithfulness of the model.



\begin{figure}[htbp]
  \centering
  \includegraphics[width=\linewidth]{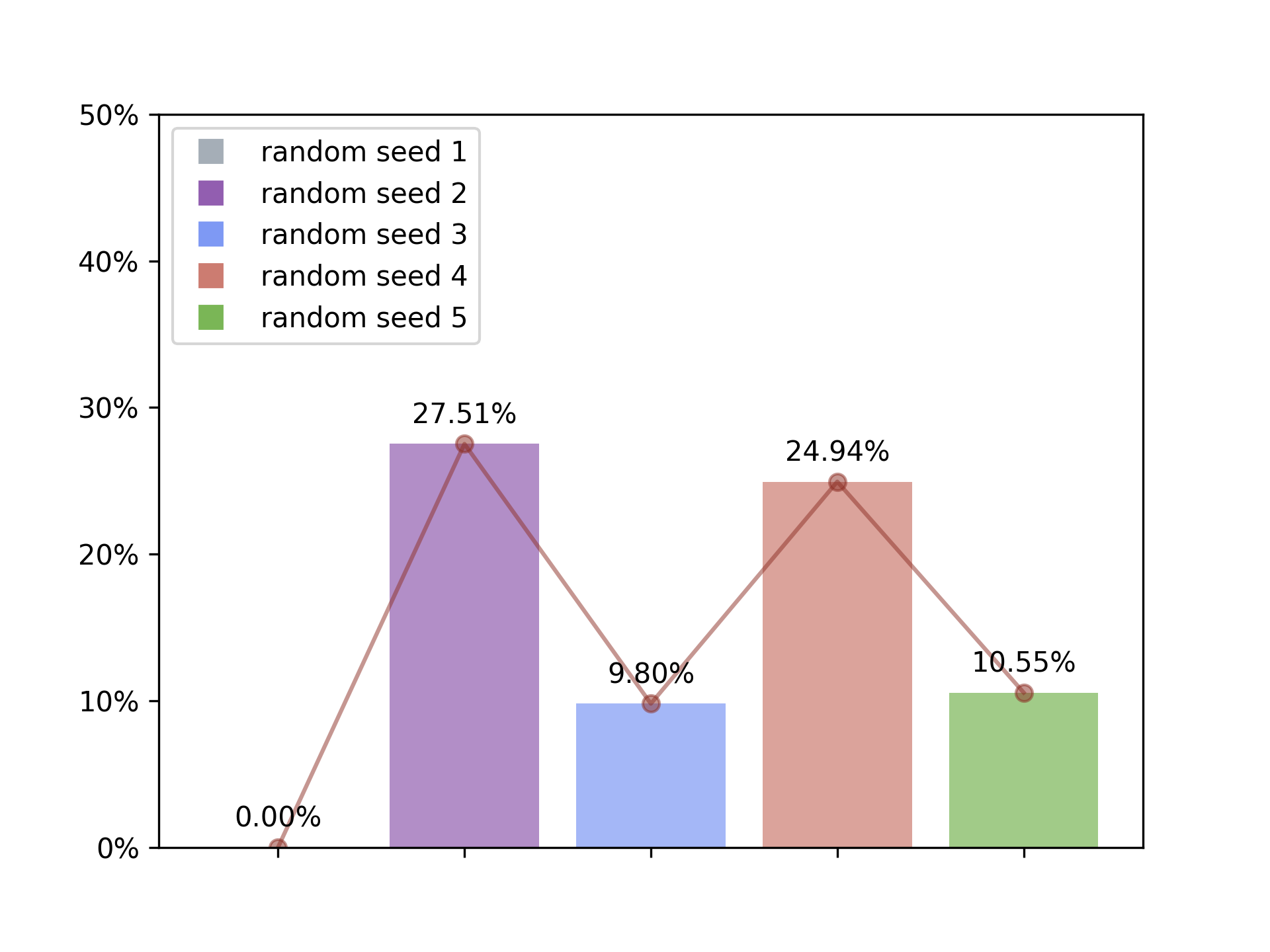}
    \vspace{-20pt}
  \caption{Five random seeds [2021,2022,2023,2024,2025] are selected to train TimeSieve respectively. With "2021" as the base model, the percentage of variation is calculated and showed.}
   \vspace{-10pt}
  \label{fig:rs_first}
\end{figure}



To address the issue of faithfulness in TimeSieve, a precise definition is required: what constitutes a \textbf{Faithful TimeSieve (FTS)} We propose that a (FTS) should embody the following three attributes:

\underline{(i)} Similarity in IB Space. 

\underline{(ii)} Closeness of Forecasting.


\underline{(iii)} Stability of Forecasting.

Our study makes the following key contributions:

(1) \textbf{Faithfulness Analysis of TimeSieve:} We conduct a comprehensive analysis to identify and understand the challenges related to the faithfulness of the TimeSieve model.

(2) \textbf{Definition of FTS:} We propose a rigorous definition for aFTS model, specifying the essential attributes required to ensure its robustness and stability.

(3) \textbf{Framework for Enhancing Faithfulness:} We develop a novel framework aimed at improving the faithfulness of TimeSieve. This framework integrates strategies to address the identified challenges while maintaining the model's performance.

(4) \textbf{Validation:} Through experimentation, we demonstrate the effectiveness of our framework in enhancing the faithfulness of TimeSieve, thereby validating its practical utility and theoretical soundness.



\section{Related Work}

\subsection{Time Series Forecasting}

Over the past few years, the domain of time series forecasting has witnessed significant progress, giving birth to a multitude of effective forecasting models. These innovative models often harness the power of Multilayer Perceptrons (MLPs), such as the MSD-Mixer \cite{zhong2024multiscale}, DLinear \cite{zeng2022transformers}, and FreTs \cite{yi2023frequencydomain}, which leverage intricate data manipulations and learning strategies. Concurrently, Transformer architectures have also gained prominence, with models like PatchTST \cite{nie2023time}, and the memory-efficient Informer \cite{zhou2021informer}, which excel in capturing temporal dependencies through advanced data transformation techniques.

In our recent work - TimeSieve \cite{TimeSieve_repo}, we have introduced an innovative approach that integrates wavelet transform with contemporary machine learning methodologies for time series forecasting. The wavelet transform initially preprocesses the input data, decomposing it into detail and trend components across various scales. This novel preprocessing step enables the model to effectively discern local variations and anomalies while seamlessly handling multi-scale information, all without introducing additional parameters, thus enhancing both model efficiency and flexibility.

\subsection{Faithful Time Series}

The faithfulness of time series predictions has been a long-standing challenge, with prior efforts focusing on developing more robust conventional methods and improving data quality. The introduction of deep learning has brought new possibilities but also introduced sensitivity to outliers and overfitting. Researchers have addressed these issues through model training enhancements, loss function modifications, ensemble methods, and data preprocessing techniques like denoising, augmentation, and decomposition. Recent works, such as DARF \cite{cheng2023weakly}, which employs adversarial learning for multi-series correlation, RNN with LSS minimization \cite{zhang2023robust} for reduced sensitivity, TimeX \cite{queen2024encoding} with model behavior consistency, and RobFormer \cite{yu2024robformer} using a Transformer-based decomposition approach, have shown promise in enhancing model faithfulness and robustness. These advancements aim to improve generalization and mitigate the impact of noise and outliers, contributing to the reliability of time series forecasting models.

These works show the great potential of learning-based model in the robust time series prediction. However,  these works do not have a unified framework and do not focus visual attention on the "Faithful" definition itself,  which is tricky.

\section{Method}

\subsection{Preliminaries: TimeSieve}

TimeSieve \cite{TimeSieve_repo} is a novel time series forecasting model that combines the strengths of Wavelet Transform and the Information Bottleneck principle to enhance prediction accuracy and robustness. The model's architecture is designed to effectively capture multi-scale features while filtering out noise and irrelevant information.

\subsubsection{Wavelet Decomposition and Reconstruction}

The Wavelet Decomposition Block (WDB) decomposes the input time series $x(t)\in\mathbb{R}^{T\times C}$ into approximation coefficients $cA$ and detail coefficients $cD$ using wavelet transform:

\begin{equation}
cA = \int x(t) \phi(t) dt
\end{equation}
\begin{equation}
cD = \int x(t) \psi(t) dt
\end{equation}

where $\phi(t)$ and $\psi(t)$ are the scaling and wavelet functions, respectively. This decomposition allows for the extraction of both trend and high-frequency details from the data. The Wavelet Reconstruction Block (WRB) then reconstructs the time series from the processed coefficients:

\begin{equation}
\hat{x}(t) = \sum c\hat{A} \phi (t) + \sum c\hat{D} \psi (t)
\end{equation}

where $c\hat{A}$ and $c\hat{D}$ are the filtered approximation and detail coefficients.

\subsubsection{Information Filtering and Compression}
The Information Filtering and Compression Block (IFCB) employs the IB principle to filter noise and retain essential information. The objective is to minimize the mutual information $I(cI; Z)$ while maximizing $I(Z; c\hat I)$, where $cI$ represents the input coefficients, $c\hat I$ are the filtered coefficients, and $Z$ is the intermediate hidden layer. The optimization problem is formulated as:

\begin{equation}
\min \{I(cI; Z) - \beta I(Z; c \hat I)\}
\end{equation}

with $\beta$ as the trade-off parameter. The IFCB uses a deep neural network with a Gaussian distribution for $p(z|i)$ and a decoder function $D(z;\theta_d)$ to predict $c\hat I$. The loss function for training combines the original prediction loss and the IB loss:

\begin{align}
&\mathcal{L} =  \mathcal{L}_{\text{o}} +  \mathcal{L}_{\text{IB}} = \notag\\
&\mathcal{L}_{\text{o}} + D_{KL}[\mathcal{N}(\mu_z, \Sigma_z) \,||\, \mathcal{N}(0, I)] + D_{KL}[p(z) \,||\, p(z|i)]
\end{align}

where $\mathcal{L}_{\text{o}}$ is the regression error, and $\mathcal{L}_{\text{IB}}$ is the IB loss.

While contemporary SOTA model TimeSieve, have made significant strides in time series forecasting by effectively harnessing wavelet transform for multi-scale feature extraction and the IB method for noise reduction, they are not without their limitations. Despite their improved predictive accuracy, these models exhibit a certain fragility when confronted with external disturbances or unforeseen perturbations in the data. This lack of robustness and stability can lead to suboptimal performance, particularly in dynamic environments where data characteristics may shift with random seeds.

\subsection{Faithfulness issues in TimeSieve}

The state-of-the-art TimeSieve model, despite its predictive prowess, displays vulnerabilities in the form of sensitivity to random seeds and input perturbations, leading to inconsistent performance. These limitations question its faithfulness and suitability for real-world scenarios where robustness is crucial. Our focus is on addressing TimeSieve's instability as a representative case, with the goal of improving trustworthiness in time series forecasting models.

By targeting these challenges, we aim to generalize our proposed solution, extending its applicability beyond TimeSieve to a broader spectrum of models. This endeavor seeks to contribute to the advancement of more reliable and robust forecasting methods, enhancing the overall confidence in time series predictions for various applications.

\subsection{What is a ``faithful time series forecasting"?} 

A ``faithful time series forecasting"  denotes a model that consistently captures the data's dynamics, delivering accurate predictions across diverse scenarios. It is robust to input perturbations, maintains performance over time, and is insensitive to initialization changes. A faithful model's stable predictions and coherent explanations foster trust, making them reliable for real-world applications and enhancing our understanding of complex temporal patterns \cite{lai2023faithful}.

\subsection{Defination for FTS}





\begin{figure*}[htbp]
\vspace{10pt}
  \centering
  \includegraphics[width=\linewidth]{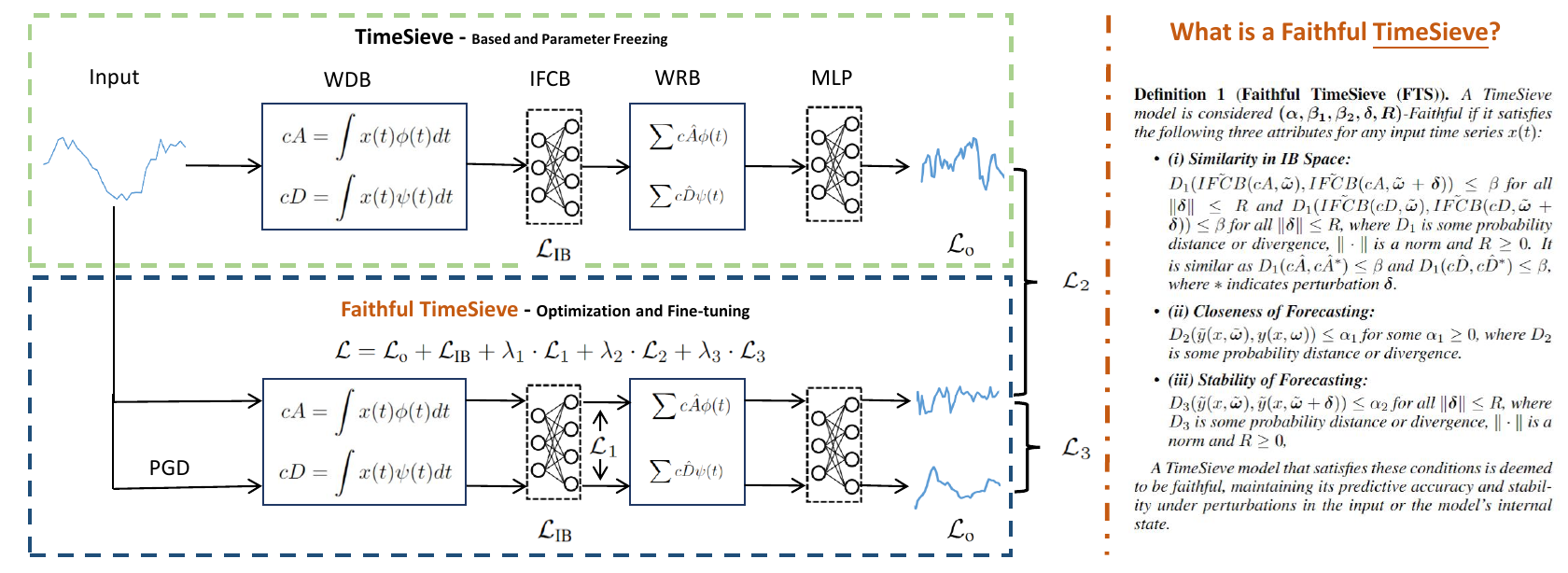}
  \caption{Framework of our proposed Faithful TimeSieve (FTS)}
  \label{fig:model}
  \vspace{-10pt}
\end{figure*}

\begin{definition}[\textbf{Faithful TimeSieve (FTS)}]\label{def:2}
A TimeSieve model is considered $\bm{(\alpha, \beta_1,\beta_2,\delta,R)}$-Faithful if it satisfies the following three attributes for any input time series $x(t)$:
\begin{itemize}
    \item \textbf{(i) Similarity in IB Space:} 

    $D_1(\Tilde{IFCB}(cA, \Tilde{\bm{\omega}}), \Tilde{IFCB}(cA, \Tilde{\bm{\omega}}+\bm{\delta})) \leq \beta$ for all $\|\bm{\delta}\| \leq R$ and $D_1(\Tilde{IFCB}(cD, \Tilde{\bm{\omega}}), \Tilde{IFCB}(cD, \Tilde{\bm{\omega}}+\bm{\delta})) \leq \beta$ for all $\|\bm{\delta}\| \leq R$, where $D_1$ is some probability distance or divergence, $\|\cdot\|$ is a norm and $R \geq 0$. It is similar as $D_1(c\hat{A},c\hat{A}^{*}) \leq \beta$ and $D_1(c\hat{D},c\hat{D}^{*}) \leq \beta$, where $*$ indicates perturbation $\bm{\delta}$.

    \item \textbf{(ii) Closeness of Forecasting:} 
    
    $D_2(\Tilde{y}(x, \Tilde{\bm{\omega}}), y(x,\bm{\omega})) \leq \alpha_1$  for some $\alpha_1 \geq 0$, where $D_2$ is some probability distance or divergence.
    
    \item \textbf{(iii) Stability of Forecasting:} 
    
    $D_3(\Tilde{y}(x, \Tilde{\bm{\omega}}), \Tilde{y}(x, \Tilde{\bm{\omega}}+\bm{\delta})) \leq \alpha_2$ for all $\|\bm{\delta}\| \leq R$, where $D_3$ is some probability distance or divergence, $\|\cdot\|$ is a norm and $R \geq 0$, 
    
\end{itemize}

A TimeSieve model that satisfies these conditions is deemed to be faithful, maintaining its predictive accuracy and stability under perturbations in the input or the model's internal state, inspired by \cite{lai2023faithful}.

\end{definition}

In the context of TimeSieve, the Wavelet Decomposition Block (WDB) and Information Filtering and Compression Block (IFCB) work together to extract and compress information effectively. The FTSe definition ensures that the model's performance is robust to variations, maintaining its faithfulness to the original time series data while providing accurate and stable forecasts.


 $D(\Tilde{y}(x, \Tilde{\bm{\omega}}), y(x,\bm{\omega})) \leq \alpha_1$  for some $\alpha_1 \geq 0$, where $D$ is some probability distance or divergence.




\textbf{Forecasting Consistency in TimeSieve.}
The similarity between forecasts from original and fine-tuned coefficients is quantified by $\alpha_1$, measuring the distance $D$ between $\Tilde{y}(x, \Tilde{\bm{\omega}})$ and $y(x,\bm{\omega})$. An ideal scenario is $\alpha_1=0$, indicating identical forecasts. Our goal is to minimize $\alpha_1$ for high forecast consistency.

\textbf{Stability in TimeSieve.}
The stability criterion is defined by $R$ and $\alpha_2$, with $R$ being the robustness radius and $\alpha_2$ the stability level. A model is highly stable if $R=\infty$ and $\alpha_2=0$, indicating immunity to perturbations. Practically, we seek large $R$ and small $\alpha_2$ for robustness.

In essence, Definition \ref{def:2} offers a holistic framework for FTS, encompassing forecast similarity, stability, and accuracy in time series forecasting.

\subsection{Faithful TimeSieve Framework}


We have already presented a rigorous definition of FTS. To construct FTS, we formulate a minimax optimization problem incorporating three conditions as outlined in Definition \ref{def:2}. The definition enables us to establish an initial optimization problem, from which we derive the following comprehensive objective function (the framework is shown in Figure \ref{fig:model}):


\begin{align}
& \underset{\|\boldsymbol{\delta}\| \leq R}{\max} \lambda_1 (\beta_1 - D_1(\Tilde{IFCB}(cA, \Tilde{\bm{\omega}}), \Tilde{IFCB}(cA, \Tilde{\bm{\omega}}+\bm{\delta}))) \nonumber \\
& + \underset{\|\boldsymbol{\delta}\| \leq R}{\max} \lambda_1 (\beta_1 - D_1(\Tilde{IFCB}(cD, \Tilde{\bm{\omega}}), \Tilde{IFCB}(cD, \Tilde{\bm{\omega}}+\bm{\delta})))  \\
& + \underset{\tilde{\omega}}{\min} \mathbb{E}_x [ \lambda_2(D_2(\tilde{y}(x, \tilde{\omega}), y(x, \omega))-\alpha_1) \nonumber \\
& + \lambda_3 ( \underset{\|\boldsymbol{\delta}\| \leq R}{\max} D_{3}(\tilde{y}(x,\tilde{\omega}), \tilde{y}(x,\tilde{\omega}+\boldsymbol{\delta})) - \beta_2 ) ] \nonumber
\end{align}

The min-max optimization problem discussed involves hyperparameters $\lambda_i$, where $i \in [3]$. 


Inspired by the Projected Gradient Descent (PGD) methodology proposed by Madry et al. (2018), the optimization process involves iterative updates to $\bm{\delta}$ and $\bm{\rho}$. At the $p$-th iteration for updating the current noise $\bm{\delta^*_{p-1}}$, we perform the following steps:

\begin{align}
\bm{\delta_p} &= \bm{\delta^*_{p-1}} +  \frac{\gamma_p }{|A_{p-1}|}\sum_{x\in A_{p-1}} \nabla_{\bm{\delta^*_{p-1}}} \nonumber\\
& [D_1(\Tilde{IFCB}(cA, \Tilde{\bm{\omega}}), \Tilde{IFCB}(cA, \Tilde{\bm{\omega}}+\bm{\delta^*_{p-1}})) \\
&+ D_2(\Tilde{IFCB}(cA, \Tilde{\bm{\omega}}), \Tilde{IFCB}(cA, \Tilde{\bm{\omega}}+\bm{\delta^*_{p-1}})) \nonumber\\
& + D_3(y(x, \bm{\Tilde{\omega}}), y(x, \bm{\Tilde{\omega} + \delta^*_{p-1}}))] \nonumber
\end{align}

\vspace{-10pt}
\begin{equation}
\bm{\delta_p^{*}} = \arg\min_{\bm{||\delta||} \leq R} ||\bm{\delta - \delta_p}||,
\end{equation}
where $A_{p-1}$ denotes a batch of samples, $\gamma_p$ is the step size parameter for PGD, and $R$ is the norm bound for the perturbation.

Once $\bm{\delta_P}$ is obtained after $P$ iterations, we update $\tilde{\omega}^{t-1}$ to $\tilde{\omega}^{t}$ using batched gradients. 

Finally, we have the following objective function:


\begin{align}\allowdisplaybreaks
&\underset{\tilde{\omega}}{\min} \mathbb{E}_x [ \lambda_1\underbrace{(D_1(\Tilde{IFCB}(cI, \Tilde{\bm{\omega}}), \Tilde{IFCB}(cI, \Tilde{\bm{\omega}}+\bm{\delta}))}_{\mathcal{L}_1,I \in [A,D]}) \nonumber\\
&+ \lambda_2\underbrace{D_2(\tilde{y}(x, \tilde{\omega}), y(x, \omega))}_{\mathcal{L}_2} \\
&+ \lambda_3\underbrace{ (  D_{3}(\tilde{y}(x,\tilde{\omega}), \tilde{y}(x,\tilde{\omega}+\boldsymbol{\delta})) }_{\mathcal{L}_3}   \nonumber ]
\end{align}

We incorporate the three mentioned losses as auxiliary attention stability losses into the original TimeSieve model loss $\mathcal{L} =  \mathcal{L}_{\text{o}} +  \mathcal{L}_{\text{IB}} $ for fine-tuning. Eventually, we obtain:
\begin{align}\mathcal{L} = \mathcal{L}_{\text{o}} +  \mathcal{L}_{\text{IB}} + \lambda_1 \cdot \mathcal{L}_1 + \lambda_2 \cdot \mathcal{L}_2 + \lambda_3 \cdot \mathcal{L}_3
\end{align}

\section{Eeperiments}

\subsection{Datasets}
In our experiment, we utilized a dataset called "Exchange Rate" \cite{lai2018modeling} which comprises the daily exchange rates of eight foreign countries over a period spanning from 1990 to 2016. The dataset includes exchange rate information for the following countries: Australia, Britain, Canada, Switzerland, China, Japan, New Zealand, and Singapore. Each entry in the dataset represents the daily exchange rate, measured against the US dollar, providing a comprehensive view of how different currencies have performed relative to the US dollar over time.


\subsection{Settings}
\begin{figure}[tbp]
\vspace{10pt}
  \centering
  \includegraphics[width=\linewidth]{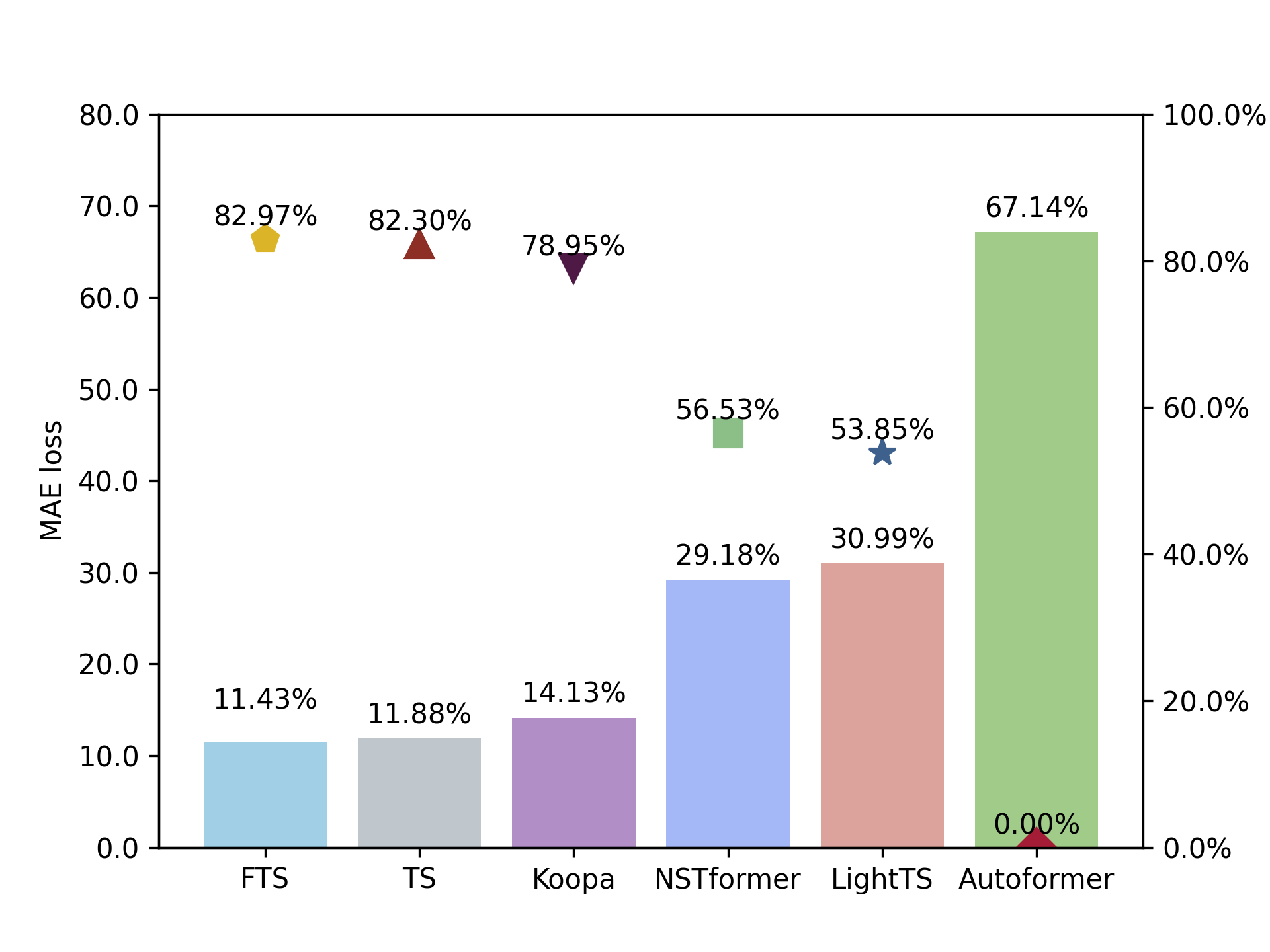}
    \vspace{-10pt}
  \caption{This figure presents a performance comparison of various advanced models, under conditions without any added perturbations. The left ordinate represents the MAE loss of the model, and the right ordinate represents the percentage of improved performance of the model over Autoformer.}
   \vspace{-10pt}
  \label{fig:sota_time_4}
\end{figure}

\textbf{Random seeds.} In the context of our randomized seed perturbation experiment, we deliberately varied the initial random seed values to assess the impact on model training and subsequent performance. Specifically, we opted for a set of five distinct seeds: 2021, 2022, 2023, 2024, and 2025 and use 2021 be the based random seed. By utilizing these different seeds, we aimed to investigate the sensitivity of the trained models to variations in the random initialization process.

\noindent \textbf{Input Perturbation. }To further evaluate the robustness of the models against perturbations, we conducted a comparative analysis. In this experiment, we introduced Gaussian noise, which is characterized by a continuous probability distribution function. Specifically, we choose the embedding $x(t)$ of the last layer of the text encoding layer and then embbed $x(t)'=x(t) + \mathcal{N}(0,\sigma)$ with a perturbation of a certain radius $\sigma=0.1$. The incorporation of Gaussian noise allowed us to simulate perturbations commonly encountered in practical scenarios and assess the models' ability to handle such disturbances.

\noindent \textbf{Setup.} The hardware utilized for our experiments comprised an NVIDIA GeForce RTX 3090 GPU and an Intel(R) Xeon(R) E5-2686 v4 CPU, ensuring the computational efficiency required for intensive model training and evaluation. We trained our models over 10 epochs with a batch size of 32 and a learning rate of 0.0001 to optimize convergence without overfitting.

To simulate adversarial conditions, we employed the Projected Gradient Descent (PGD) attack with parameters set to epsilon=0.1, alpha=1/255, and a step count of 10. Our experimental design also included a look-back window of 288 and a prediction window of 144.

\subsection{Results}
\textbf{Performance.}

We first compare FTS with other state-of-the-art models (TS \cite{TimeSieve_repo}, Koopa \cite{liu2024koopa}, Non-stationary Transformers(NSTformer) \cite{liu2023nonstationary}, LightTS \cite{campos2023lightts},  Autoformer \cite{wu2021autoformer})), and the results are shown in Figure \ref{fig:sota_time_4}. 
We are surprised to find that our framework not only did not perform worse in the original case without additional perturbations, but instead managed to achieve a higher performance (SOTA). The explanation for our preliminary analysis is that the data for timing problems inherently has perturbations, and our method effectively improves the generalization of the model, making it perform better on the original task. 

\begin{figure}[tbp]
\vspace{10pt}
  \centering
  \includegraphics[width=\linewidth]{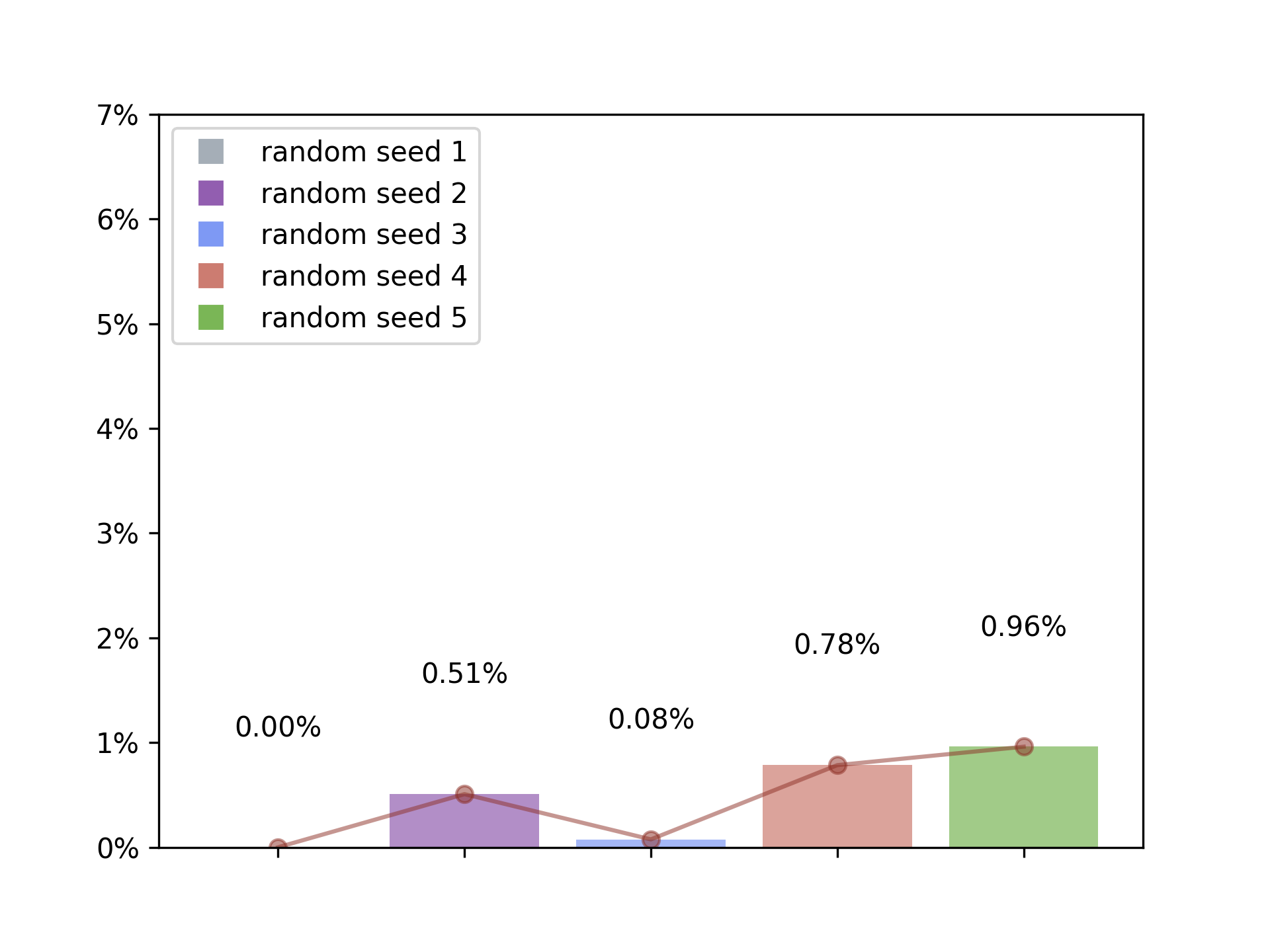}
    \vspace{-10pt}
  \caption{Five random seeds [2021,2022,2023,2024,2025] are selected to train Faithful TimeSieve respectively. With "2021" as the base model, the percentage of variation is calculated and showed.}
   \vspace{-10pt}
  \label{fig:rs}
\end{figure}

The left ordinate represents the MAE loss of the model, while the right ordinate represents the percentage of improved performance of the model over Autoformer. It can be seen that even with only Faithful optimization, the performance on the original dataset without added noise is still superior to that of the unoptimized model.


\begin{table}[tbp]
  \centering
  \caption{Performance comparison of FTS and TS under different random seeds shows that FTS performs better than TS and is less affected by the choice of random seed.}
    \begin{tabular}{c|ccc}
    \toprule
    \multicolumn{1}{c}{\textbf{Random seed}} & \textbf{TS}    & \textbf{FTS}   & \textbf{Preference(\%)} \\
    \midrule
    \underline{2021*}  & 0.1639 & 0.1155 & \textcolor{blue}{NA} \\
    2022  & 0.1188 & 0.1148 & \textcolor{blue}{\textbf{98.16\%}} \\
    2023  & 0.1479 & 0.1155 & \textcolor{blue}{\textbf{99.23\%}} \\
    2024  & 0.1230 & 0.1146 & \textcolor{blue}{\textbf{96.86\%}} \\
    2025  & 0.1466 & 0.1143 & \textcolor{blue}{\textbf{90.90\%}} \\
    \bottomrule
    \end{tabular}%
  \label{tab:per_FTS_FS_seed}%
\end{table}%


\noindent \textbf{Faithfulness. }To demonstrate the stability of our model's performance across various random seeds, we conducted a comparative analysis of the performances of FTS and TS under multiple seed settings, with detailed results presented in Table \ref{tab:per_FTS_FS_seed}. We specifically chose the performance metrics from the 2021 seed as our based model, which allowed us to quantitatively assess how FTS diminishes the variability induced by different seeds compared to TS. This analysis revealed that FTS consistently reduced the influence of seed variability on performance metrics, with a significant reduction of up to 99.23\%.

We selected the performance under the random seed 2021 as our based model and calculated the reduction in seed variability by FTS compared to TS. Figures  \ref{fig:rs_first} and Figures \ref{fig:rs} clearly illustrate that FTS maintains more consistent performance than TS.

\begin{table}[htbp]
  \centering 
  \caption{In experiments comparing noise addition to the test set under a fixed random seed, the results demonstrate that FTS exhibits superior robustness compared to TS.}
    \begin{tabular}{c|cc}
    \toprule
    \multicolumn{1}{c}{\textbf{Setting}} & \textbf{TS}    & \textbf{FTS} \\
    \midrule
    \textbf{Non-perturbation} & 0.1188  & 0.1149  \\
    \textbf{$\mathcal{N}(0,\sigma)$} & 0.1525  & 0.1203  \\
    \textbf{Preference(\%)} & 28.35\% & \textcolor{red}{\textbf{4.71\%}} \\
    \bottomrule
    \end{tabular}%
  \label{tab:rob}%
\end{table}%

To demonstrate the robustness of the FTS, we conducted input noise experiments using the random seed 2022 to add noise to the test set. Under these conditions, the MAE loss of the TS increased from 0.1188 to 0.1554, whereas the MAE loss of the FTS model only slightly deteriorated from 0.1148 to 0.1202. This resulted in a significant reduction in the impact of input noise from 28.4\% to 4.7\%, greatly diminishing the disturbance caused by noise. The specific experimental results, as shown in Table \ref{tab:rob}, indicate that our optimized FTS model exhibits substantial robustness compared to the ST model.


\section{Conclusions}

In conclusion, our study has illuminated the critical need for faithfulness in time series forecasting models, particularly in the context of TimeSieve. The identification of unfaithfulness issues, such as high sensitivity to random seeds and input noise, has underscored the importance of developing robust and reliable forecasting mechanisms. In response, we have proposed a novel framework tailored to enhance the faithfulness of TimeSieve by a rigorous definition which we propose, mitigating the effects of these vulnerabilities.

Through demo level experimentation, we have demonstrated the effectiveness of our proposed framework in improving the stability and resilience of TimeSieve's predictions. The results not only validate our approach but also highlight its potential to foster a new standard in the development of time series forecasting models.

We're fully aware that there's a lot more to discover about this framework, but we wanted to get our amazing findings out there right away. As a future course of action, we intend to do more comprehensive and perfect experiments to find more interesting things and broaden our research by extending the evaluation of our framework to a more diverse set of scenarios and datasets. This will further solidify the generalizability and applicability of our method. Ultimately, we envision our work serving as a foundation for enhancing the faithfulness of not just TimeSieve but also other advanced temporal models, thereby contributing to the overall advancement and reliability of time series forecasting in various domains and industries. Our efforts aim to strengthen the trustworthiness of these models, ultimately benefiting decision-making processes that rely on accurate and consistent predictions.


\section*{Ethical Statement}

There are no ethical issues.



\bibliographystyle{named}
\bibliography{ijcai24}

\end{document}